# RetinaVision: XAI-Driven Augmented Regulation for Precise Retinal Disease Classification using deep learning framework

Mohammad Tahmid Noor
*Dept. of CSE*
*East West University*
Dhaka, Bangladesh
tahmidnoor770@gmail.com

Shayan Abrar
*Dept. of CSE*
*American International University-Bangladesh*
Dhaka, Bangladesh
shayanabrar7@gmail.com

Jannatul Adan Mahi
*Dept. of CSE*
*East West University*
Dhaka, Bangladesh
jannatuladanmahi@gmail.com

Md Parvez Mia
*Dept. of CSE*
*East West University*
Dhaka, Bangladesh
mdparvezmia999@gmail.com

Asaduzzaman Hridoy
*Dept. of ICT*
*Bangladesh Army University of Engineering and Technology*
Dhaka, Bangladesh
asaduzzamanhridoyice@gmail.com

Samanta Ghosh
*Dept. of CSE*
*East West University*
Dhaka, Bangladesh
Samantaewu28@gmail.com

***Abstract*— Early and accurate classification of retinal diseases is critical to counter vision loss and for guiding clinical management of retinal diseases. In this study, we proposed a deep learning method for retinal disease classification utilizing optical coherence tomography (OCT) images from the Retinal OCT Image Classification – C8 dataset (comprising 24,000 labeled images spanning eight conditions). Images were resized to 224×224 px and tested on convolutional neural network (CNN) architectures: Xception and InceptionV3. Data augmentation techniques (CutMix, MixUp) were employed to enhance model generalization. Additionally, we applied GradCAM and LIME for interpretability evaluation. We implemented this in a real-world scenario via our web application named RetinaVision. This study found that Xception was the most accurate network (95.25%), followed closely by InceptionV3 (94.82%). These results suggest that deep learning methods allow effective OCT retinal disease classification and highlight the importance of implementing accuracy and interpretability for clinical applications.**

## I. Introduction

An important part of the eye, the retina, contributes to processing visual information. Diseases of the retina, including diabetic retinopathy, age-related macular degeneration, and glaucoma, can eventually result in irreversible vision loss if left undetected and untreated. Early diagnosis using diagnostic imaging, especially Optical Coherence Tomography (OCT), has become popular as a non-invasive method for evaluation of retinal health. However, the process of manually analyzing OCT images is time-consuming and subjective due to human error, so there is a need for exploring automatic methods that can classify diseases with good accuracy.

Traditional retinal disease detection approaches, which are mainly based on clinical diagnostic and manual examination of ocular fundus images, are still limited by subjectivity and reliance on expertise. These approaches suffer from interoperator variability and misinterpretation of subtle retinal signs, and cause delay in diagnosis, which makes early detection very difficult. Further, manual analysis can be slow and may not be scalable in resource-poor settings. These constraints emphasize the critical need for more reliable, accurate, and automatic diagnostic systems, which could improve consistency, decrease time delay, and help early treatment in the sector of retinal health care.

Some authors have investigated the use of automated methods for retinal disease diagnosis. For instance, Kulyabin et al. [1] presented an open-access OCT dataset and obtained exemplary performance using deep learning models for classifying these image categories. Their work enabled benchmarking of deep learning algorithms, demonstrating that convolutional models could reliably distinguish disease categories with strong generalization. This dataset has become a valuable resource for researchers addressing the scarcity of publicly available retinal imaging data.

Similarly, Rahimzadeh et al. [2] designed an ensemble CNN model with combined EfficientNetV2-B0 and Xception networks, which achieved better spatial resolution learning for retinal disease detection and further increased the classification accuracy to a remarkable level. This ensemble design highlighted the potential of combining lightweight and deeper models for diagnostic applications. Additionally, Kim et al. [3] used multiple binary CNN classifiers on OCT images and were able to obtain high accuracy on a few diseases, including Choroidal Neovascularization (CNV), Diabetic Macular Edema (DME), and Drusen, showcasing that diseasespecific binary models can outperform multi-class classifiers. This strategy emphasized precision in handling class imbalances and improved reliability for specific retinal pathologies.

Despite great achievements, deep learning models still suffer from challenges in retinal OCT image classification due to the requirement of massive labeled datasets, high computational cost, and the complexity of noise arising from different acquisition systems and variability. This research is motivated by these challenges and the paucity of previous work that has addressed them effectively. To strengthen our approach, we propose employing Grad-CAM for model interpretability, integrating CutMix and MixUp for data augmentation, and deploying the framework in a real-world setting through a practical application. Our work prefers to establish a controlled baseline but we plan to expend our work by using multi-center OCT database in future. The goal of this paper is to establish a strong foundation for retinal disease classification by leveraging state-of-the-art deep learning techniques. Research Questions:

- How can deep learning models be optimized for more precise and faster classification of retinal OCT images?
- What methods can be implemented to overcome the limitations of current datasets and image quality in retinal OCT image analysis?

## II. RELATED WORKS

Verma [4] presents a Convolutional Neural Network (CNN) model designed for classifying four retinal diseases: normal, cataract, diabetic retinopathy, and glaucoma. The model was developed and tested on a large dataset of annotated retinal images, achieving an overall accuracy of 84%. Notably, diabetic retinopathy classification demonstrated almost perfect performance, while glaucoma classification exhibited the lowest accuracy. The study highlights the necessity for further modifications, such as regularization and data augmentation, to enhance the model's clinical applicability.

Rithani et al. [5] employ deep learning models, specifically VGG16 and InceptionV3, for the classification of retinal diseases in OCT images. Utilizing a dataset of approximately 85,000 high-resolution images, the models effectively classify primary retinal conditions such as diabetic macular edema (DME), age-related macular degeneration (AMD), and choroidal neovascularization (CNV). VGG16 demonstrates superior performance in classifying CNV, while InceptionV3 excels in identifying healthy retinal conditions, achieving an overall accuracy of 92.76% for both models.

Vasanthi et al. [6] focus on the automated classification of macula optical coherence tomography (OCT) images to differentiate various retinal diseases using deep learning techniques. It employs convolutional neural network (CNN) architectures, including ManualNet, AlexNet, and ResNet, achieving varied accuracy levels. Notably, ResNet attained the highest accuracy at 90%. The research highlights the effectiveness of these models in accurately classifying diseases such as Age-Related Macular Degeneration, Diabetic Macular Edema, and others, enhancing early detection and treatment.

Eren et al. [7] introduced a method for classifying retinal diseases in OCT images using transfer learning. By segmenting retinal layers and integrating these segmentations with OCT scans, the researchers achieved a classification accuracy of 91.47% and an Area Under the Curve (AUC) of 0.96. This approach involved combining horizontal and vertical cross-sectional middle slices with segmentation outputs, significantly enhancing the classification performance compared to traditional OCT slice images.

Wali et al. [8] employ a random forest classifier for the classification of retinal diseases using optical coherence tomography (OCT) images. It introduces a novel approach called raw image data embedding (RIDE), which utilizes raw image data instead of metadata-driven preprocessing. The model achieved an accuracy rate of 80% in classifying conditions such as diabetic macular edema, choroidal neovascularization, and DRUSEN, outperforming various other classifiers and established image processing methods, thereby enhancing diagnostic accuracy and efficiency.

Jannat et al. [9] present OCT-SelfNet, a self-supervised machine learning framework designed for retinal disease detection using optical coherence tomography (OCT) images. It employs a two-phase training approach that combines self-supervised pretraining and supervised fine-tuning with a mask autoencoder based on the SwinV2 backbone. The model demonstrates superior performance, achieving AUC-ROC scores exceeding 77% and AUC-PR scores surpassing 42%, significantly outperforming the baseline model, Resnet-50, which achieved AUC-ROC scores of over 54% and AUC-PR scores of over 33%.

## III. METHODOLOGY

Early and accurate classification of retinal diseases is critical to counter vision loss and to guide clinical management of retinal diseases. In this study, we proposed a deep learning method showcased in Fig. 1 for retinal disease classification utilizing optical coherence tomography (OCT) images from the Retinal OCT Image Classification – C8 dataset (comprising 24,000 labeled images spanning eight conditions). Images were resized to 224 × 224 px and tested on convolutional neural network (CNN) architectures: Xception and InceptionV3. Data augmentation techniques (CutMix, MixUp) were employed to enhance model generalization. Additionally, we applied Grad-CAM and LIME for interpretability evaluation. This study found that Xception was the most accurate network (95%), followed closely by InceptionV3 (94%). These results suggest that deep learning methods allow effective OCT retinal disease classification and highlight the importance of implementing accuracy and interpretability for clinical application.

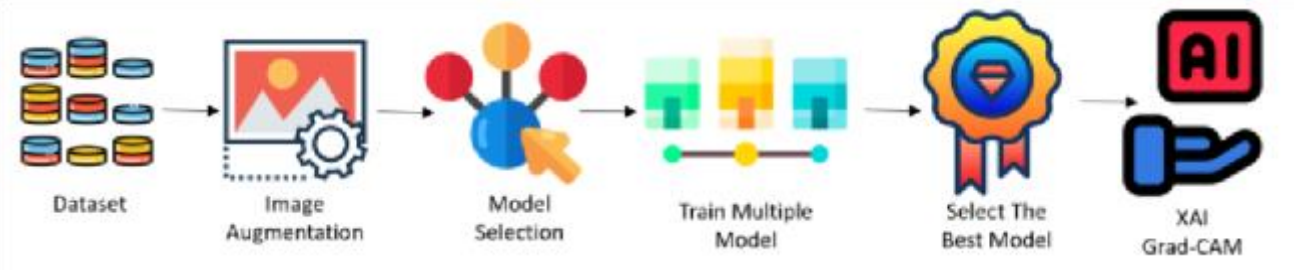

Fig. 1. Proposed Methodology

The Retinal OCT Image Classification – C8 dataset is a broad and high-quality dataset of optical coherence tomography (OCT) images made to support research on the classification of retinal diseases demonstrated in Fig. 2.

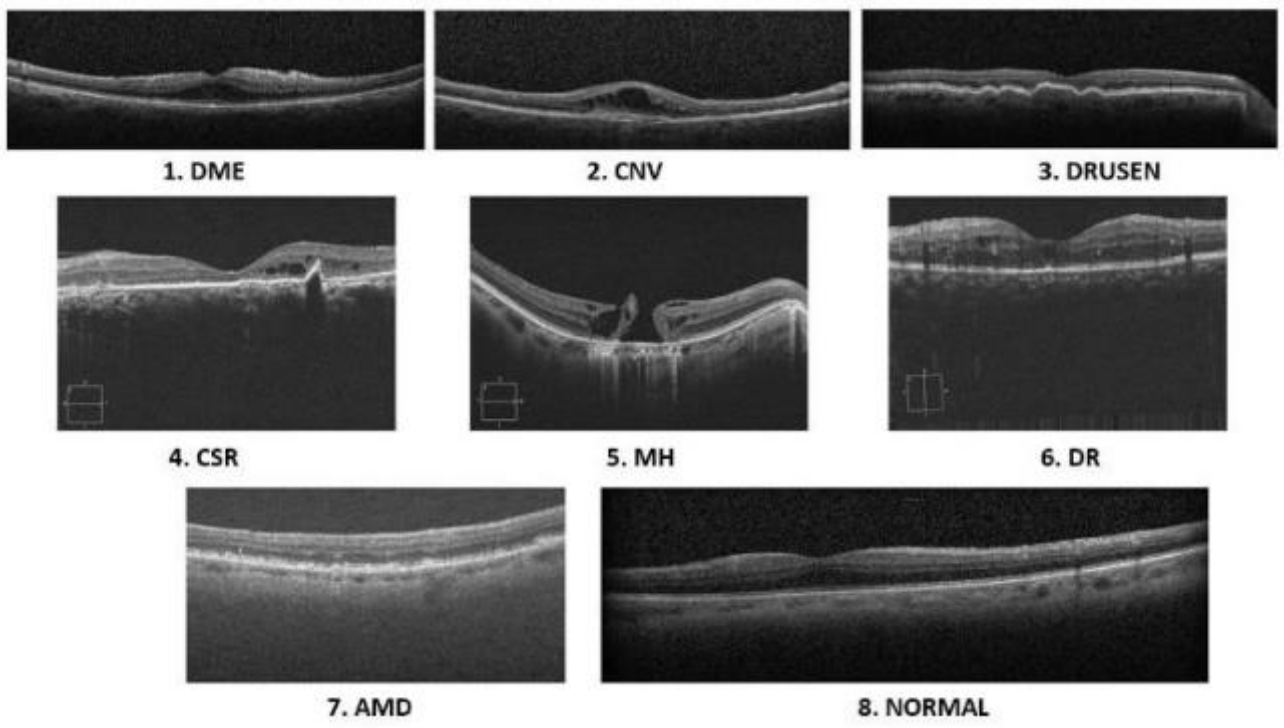

Fig. 2. Dataset Sample

It contains 24,000 labeled images present in eight retinal disease categories, which include retinal diseases such as diabetic retinopathy, age-related macular degeneration (AMD), and clinically relevant retinal diseases. First one is DME, second one is CNV, third one is DRUSEN, fourth one is CSR, 5th is MH, sixth one is DR, 7th is AMD, and the last one is a NORMAL dataset image. The class distribution in our dataset is sufficiently representative. Each image included in the dataset comes from an OCT scan, which provides a detailed cross-sectional view of retinal structures needed to identify retinal disease accurately

### A. *Xception*

Utilizing the idea of depth-wise separable convolutions, the Xception architecture shown in Fig. 3 is a deep convolutional neural network, as depicted in the diagram. Entry flow, Middle flow, and Exit flow are its three main elements of it [9]. The input image is first handled by the entry flow using a sequence of 3x3 convolutions, which gradually decreases the spatial dimensions while increasing the number of feature maps. Multiple repeated blocks of 3x3 convolutions make up the middle flow, which generates a feature map that captures intricate spatial hierarchies and attributes. After implementing additional convolutions, the exit flow uses global average pooling to shrink the spatial dimensions before reducing via a softmax activation for classification.

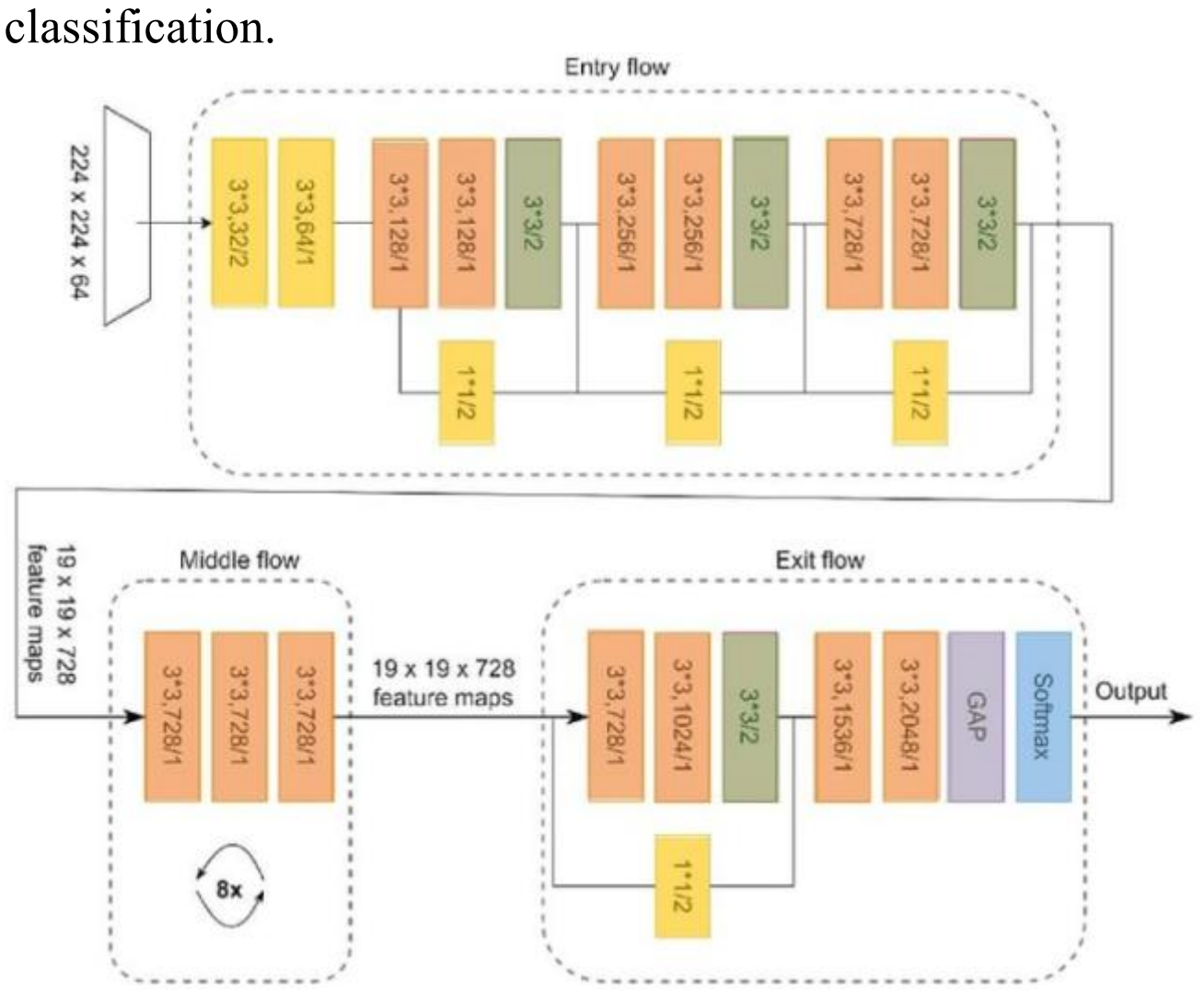

Fig. 3. Proposed Xception Architecture

### B. *InceptionV3*

The flow of an image through a convolutional neural network for classification appears in the illustration of Fig 4.

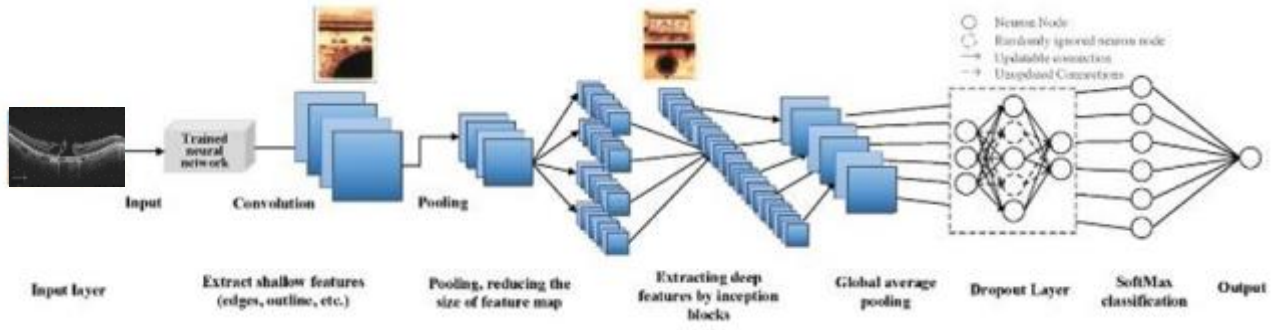

Fig. 4. Proposed InceptionV3 Architecture

The input layer, where the image is put into the network, is where the process starts. Convolution is the initial phase in which superficial features like edges and outlines are removed. To minimize computational complexity, these features are run through a pooling layer, which cuts the feature maps' size. Inception blocks then learn complex patterns to detect deeper features. Global average pooling serves to further compress the feature maps, ending in a compact representation. Better generalization is guaranteed by applying a dropout layer to avoid overfitting. Lastly, a SoftMax layer that calculates classification probabilities is utilized to classify the output [10].

## IV. AUGMENTED REGULATION

### A. *CutMix*

A patch from a different image in the CutMix is used to replace a portion of one image while the image is being adjusted accordingly, as visually illustrated in Fig. 5. The sum is weighted according to the patch area. CutMix is described as follows for the input and label represented by T and Q:

$$T_new = M \odot T_i + (1-M) \odot T_j$$
$$Q_new = \partial Q_i + (1-\partial)\, Q_j$$

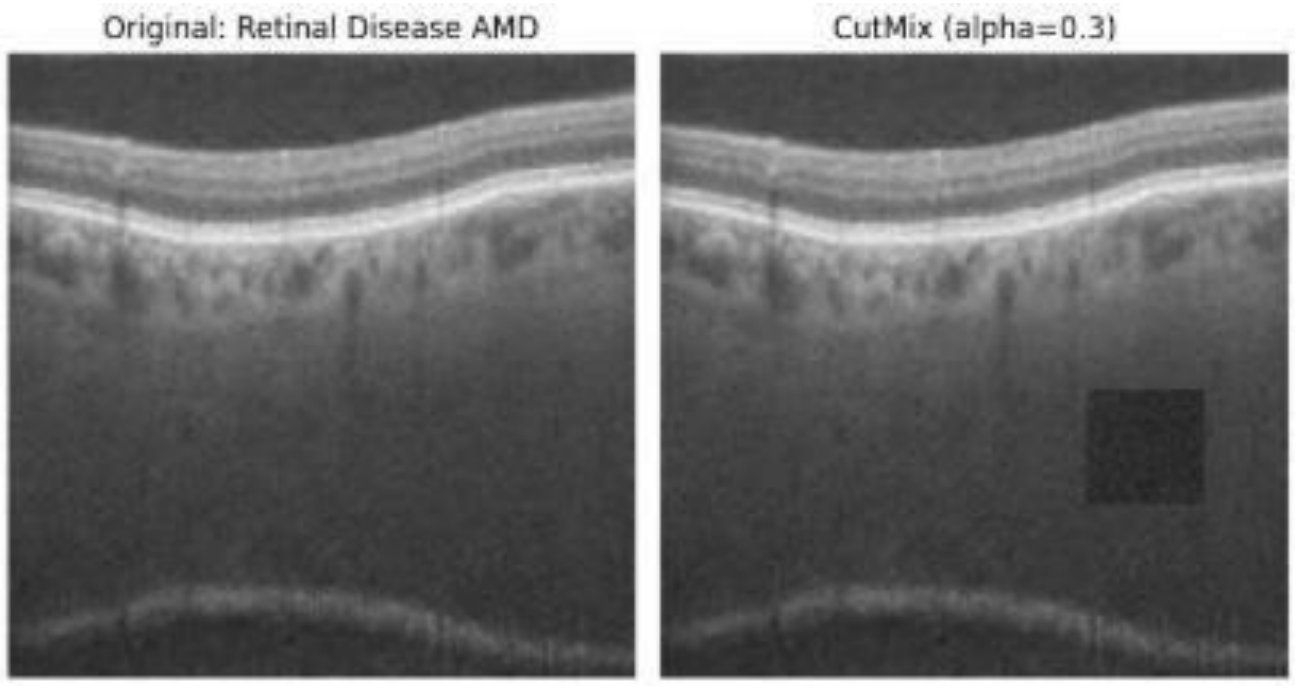

Fig. 5. CutMix image for AMD disease, alpha = 0.3

### B. *MixUp*

MixUp is the process of combining two images by linearly interpolating the labels and pixel values, as visually demonstrated in Fig. 6. The labels for the randomly chosen images p_i and p_j will be q_i and q_j, respectively. MixUp is defined as:

$$P_new = \partial pi + (1-\partial)pj$$
$$Q_new = \partial qi + (1-\partial)qj\ .$$

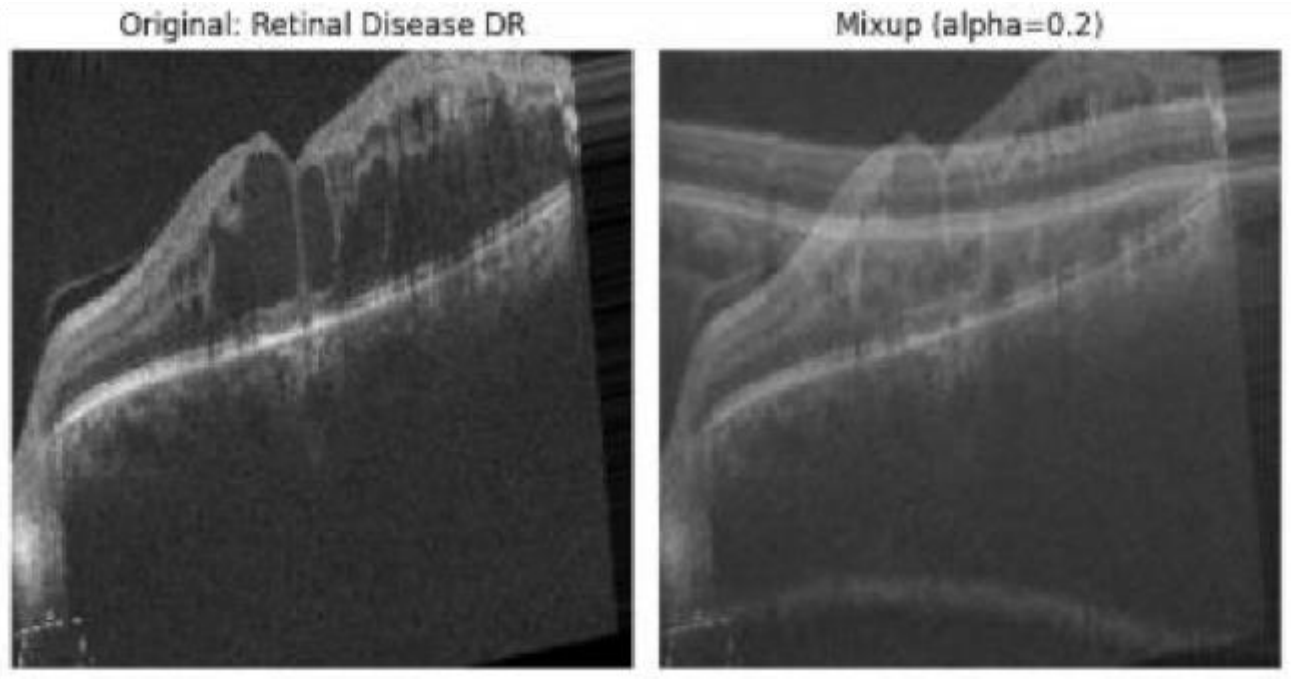

Fig. 6. MixUp image for DR disease, alpha = 0.2

## V. EXPERIMENTAL ANALYSIS

### A. *Xception*

The plot shown in Fig. 7 shows the accuracy and loss over 50 epochs for both training and validation.

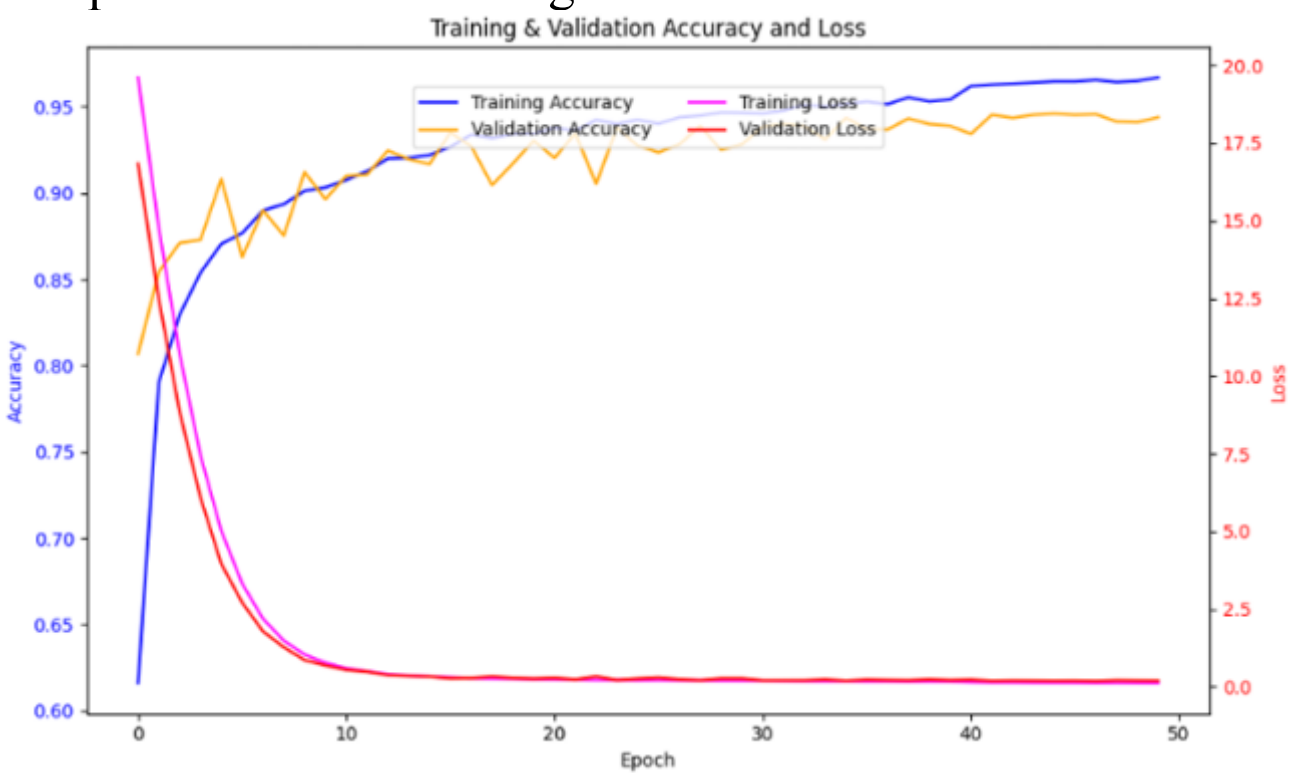

Fig. 7. Accuracy and Loss for training and validation of Xception

Validation accuracy stabilizes at 90%. While training accuracy increases rapidly to approximately 95%. The validation loss shows a similar trend but stays higher, suggesting some overfitting, whereas the training loss drops off dramatically [12]. The difference between validation and training loss drops off dramatically. The difference between validation and training losses indicated that the model overfitted at first, then improved in subsequent epochs. Despite that, the model shows good learning overall.

The classification models' efficiency across eight retinal disease classes is highlighted in the confusion matrix shown in Fig. 8. The predicted class is shown in each column, and the true class appears in each row. With the majority of values across the diagonal, the model performs well and shows precise classifications. There are certain misclassifications in the normal class, but the model shows outstanding accuracy with few errors in every class.

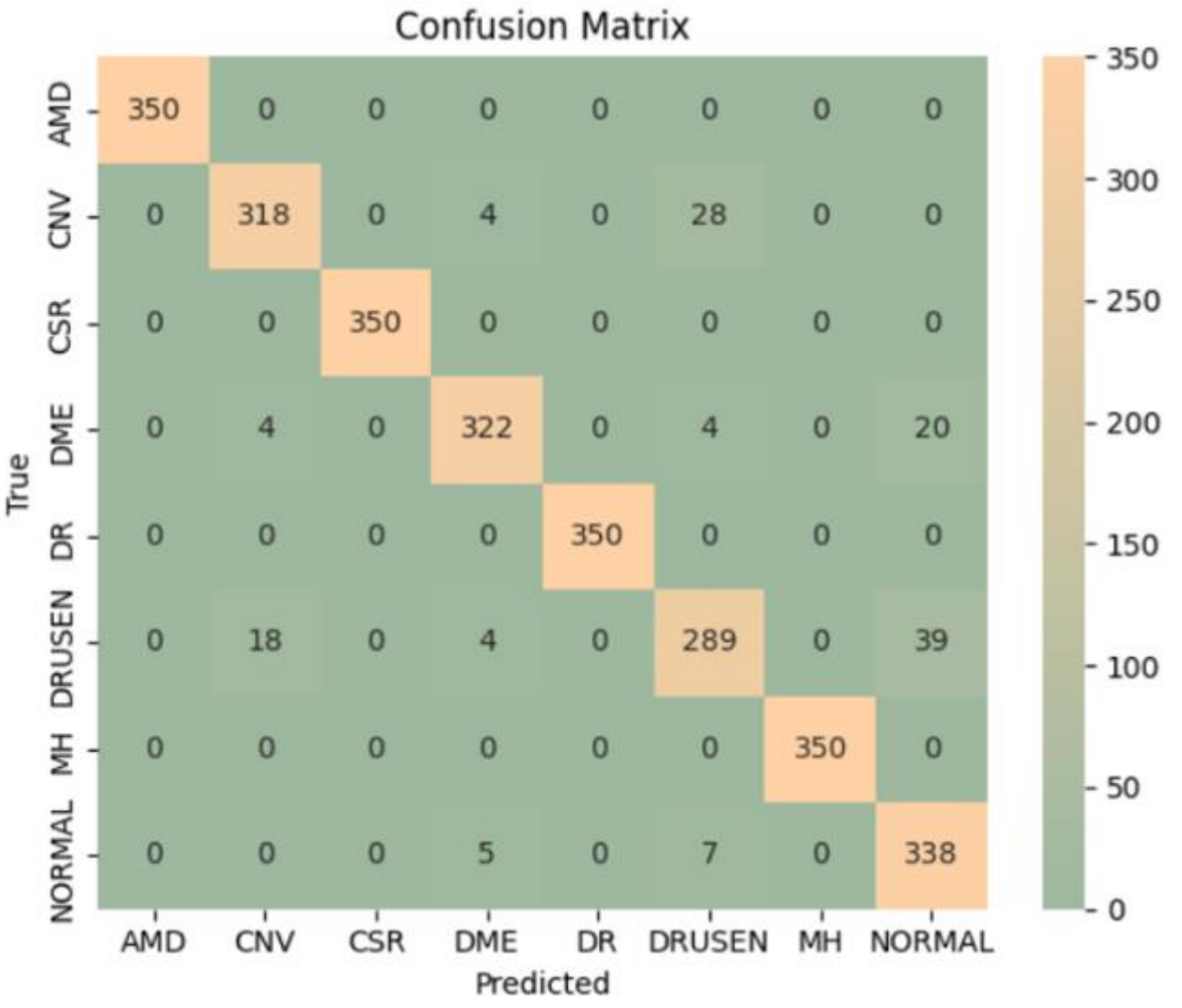

Fig. 8. Confusion Matrix of Xception

The Xception model's classification report, shown in Table 1, performs well across all retinal disease classes, according to the classification report. The model's flawless classification for AMD, CSR, DR, and MH is demonstrated by its perfect precision, recall, and F1-score.

TABLE I CASSIFICATION REPORT OF XCEPTION

| Model Name | Classes | Precision | Recall | F1-Score |
|---|---|---|---|---|
| Xception | AMD | 1.00 | 1.00 | 1.00 |
| | CNV | 0.94 | 0.91 | 0.92 |
| | DSR | 1.00 | 1.00 | 1.00 |
| | DME | 0.96 | 0.92 | 0.94 |
| | DR | 1.00 | 1.00 | 1.00 |
| | DRUSEN | 0.88 | 0.83 | 0.85 |
| | MH | 1.00 | 1.00 | 1.00 |
| | NORMAL | 0.85 | 0.97 | 0.90 |

The CNV, DME, and NORMAL classes also exhibit high values. With a F1-score of 0.85, recall of 0.83, and precision of 0.88, the DRUSEN class performs well but is marginally worse than the other class. All things considered, the model does a great job of classifying retinal diseases.

The primary hyperparameters for model training are given in the Table 2. To maintain model performance and computational efficiency, a batch size of 32 was selected. For a multi-class classification task, the categorical crossentropy loss function is utilized, which works well. To allow for gradual convergence during training, the learning rate is adjusted to 0.0001. Adam optimizer with adaptive learning rates is utilized. To avoid overfitting and ensure that the model generalizes well, training takes place over 50 epochs with early stopping initiated after 10 epochs of no improvement in validation loss.

TABLE II HYPERPARAMETER TUNING XCEPTION

| | | | |
|---|---|---|---|
| **Batch size** | 32 | **Loss function** | categorical_crossentropy |
| **Learning rate** | 0.0001 | **Number of epochs** | 50 |
| **Optimizer** | Adam | **Patience** | 10 |

## *B. InceptionV3*

The training and validation accuracy over 50 epochs is displayed in Fig. 9. The training accuracy roughly rises to approximately 95% at the conclusion. Apart from that, increasing the validation accuracy settles at roughly 90%. The validation accuracy fluctuates a little, suggesting a slight degree of overfitting early in the training, but the accuracy increases over time. The model operates well overall, indicating strong generalization and learning.

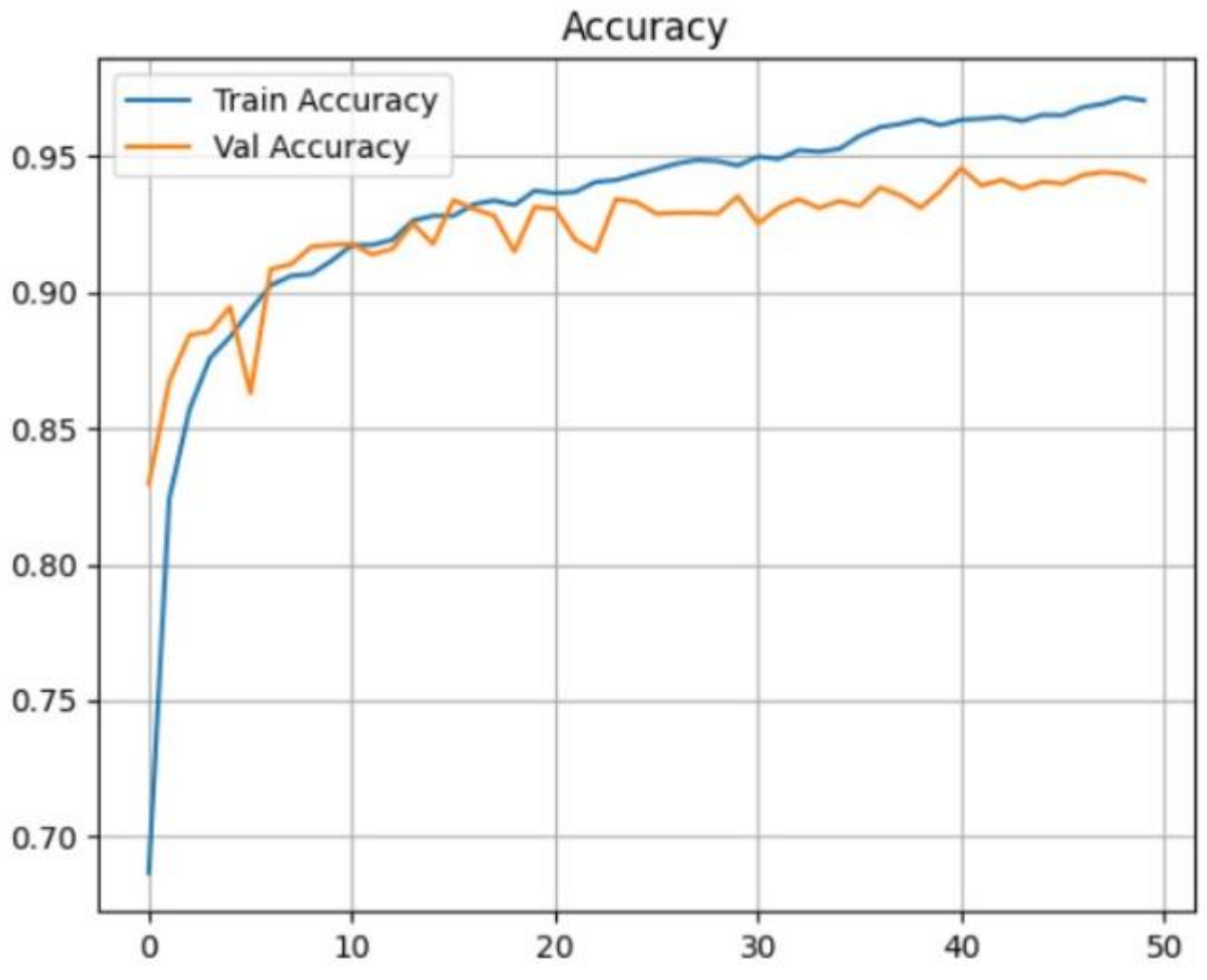

Fig. 9. Training and validation accuracy curve of InceptionV3

Over 50 epochs, the plot in Fig. 10 illustrates training and validation loss.

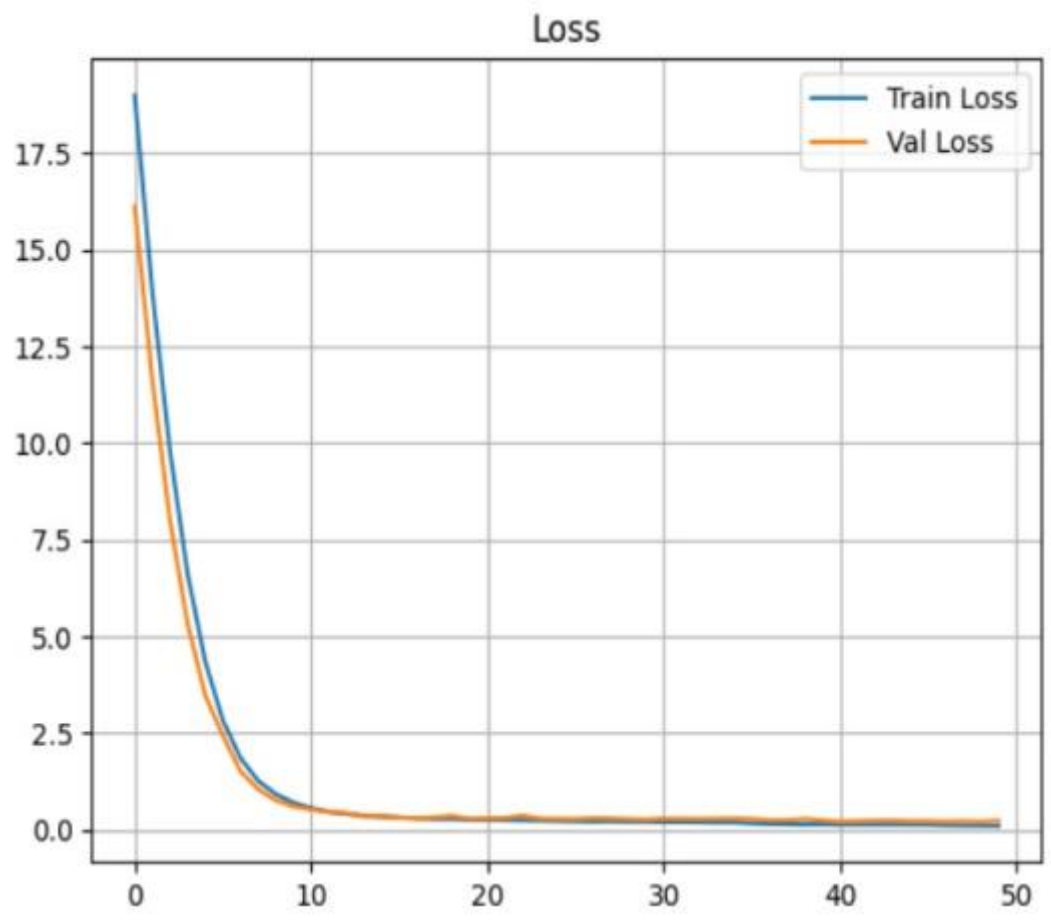

Fig. 10. Training and validation loss curve of InceptionV3

Both first reduce dramatically, demonstrating successful learning, and then level off near zero, indicating convergence. The training and validation loss curves are similar, suggesting good generalization and the lack of overfitting. The steady decline validates the efficiency of the training procedure.

The model's performance across eight retinal disease classes is displayed in the confusion matrix in Fig. 11. High values along the diagonal indicate that most predictions are accurate. AMD, for instance, has 350 accurate predictions [13]. CNV also has good classifications overall. Although NORMAL has very minor misclassifications, its accuracy remains high. Overall, the matrix shows that the model has learned efficiently.

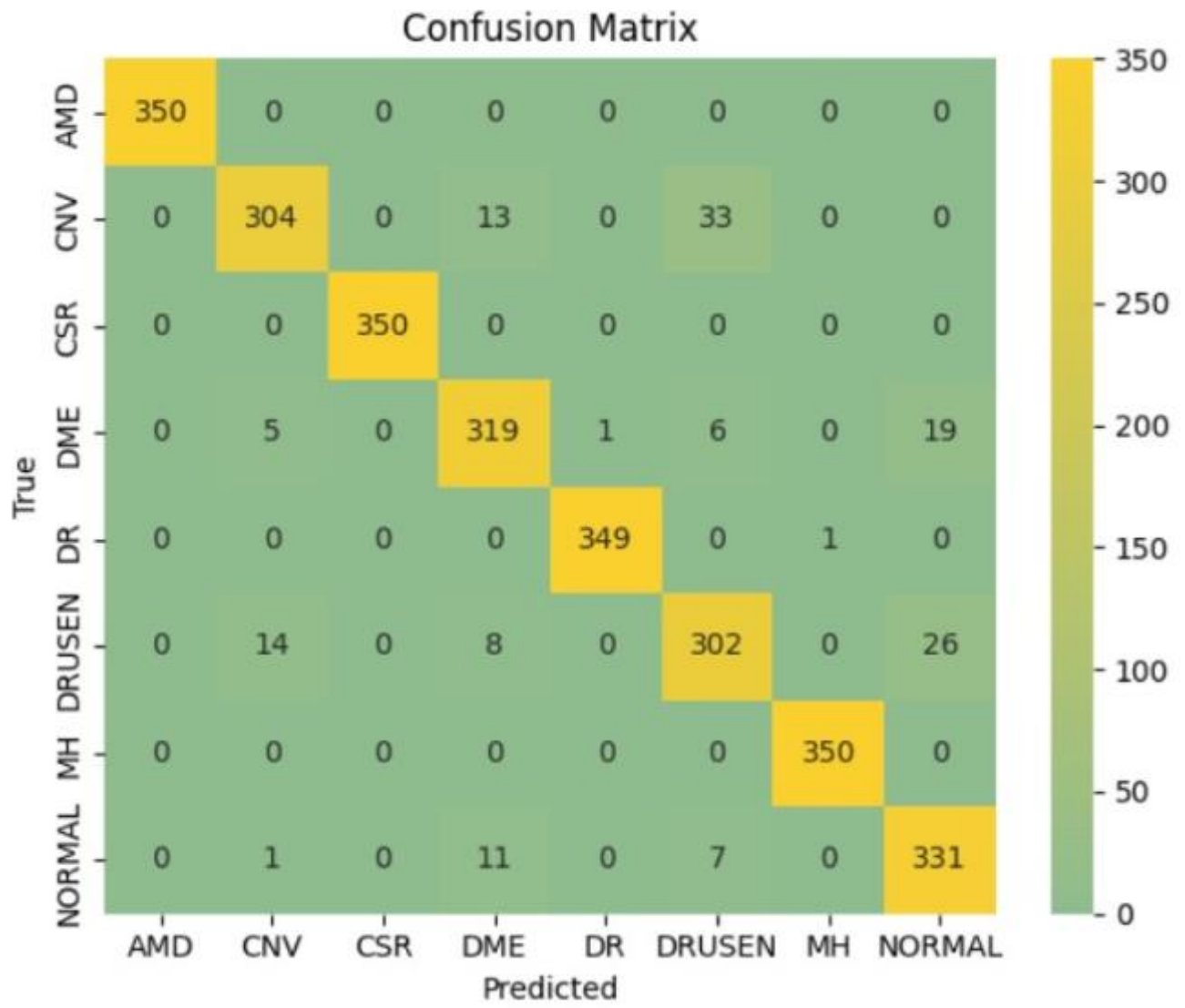

Fig. 11. Confusion Matrix of InceptionV3

Hyperparameters for training the Inceptionv3 model are outlined in Table 3. A batch size of 32 ensures stability and efficient computation. For multi-class classification, categorical cross-entropy is utilized, with a learning rate of 0.0001 to ensure stable consistency. Overfitting is prevented, and optimal performance is guaranteed with the Adam optimizer, 50 epochs, and early stopping [11].

TABLE III HYPERPARAMETER TUNING OF INCEPTIONV3

| | | | |
|---|---|---|---|
| **Batch size** | 32 | **Loss function** | categorical_crossentropy |
| **Learning rate** | 0.0001 | **Number of epochs** | 50 |
| **Optimizer** | Adam | **Patience** | 10 |

Through perfect precision, recall, and F1-score of 1.00 for AMD, CSR, DR, and MH, the Inceptionv3 model excels in the majority of retinal disease classes illustrated in Table 4. CNV operates well, with a precision of 0.94, a recall of 0.87, and an F1-score of 0.90. With a precision and recall of 0.91, DME similarly does well. Despite having F1-scores of 0.87 and 0.91, respectively, DRSEN and NORMAL perform well overall in classification.

TABLE IV CLASSIFICATION REPORT OF INCEPTION

| Model Name | Classes | Precision | Recall | F1-Score |
|---|---|---|---|---|
| InceptionV3 | AMD | 1.00 | 1.00 | 1.00 |
| | CNV | 0.94 | 0.87 | 0.90 |
| | CSR | 1.00 | 1.00 | 1.00 |
| | DME | 0.91 | 0.91 | 0.91 |
| | DR | 1.00 | 1.00 | 1.00 |
| | DRUSEN | 0.87 | 0.86 | 0.87 |
| | MH | 1.00 | 1.00 | 1.00 |
| | NORMAL | 0.88 | 0.95 | 0.91 |

## VI. Explainable AI

### *A. Grad-CAM Visualization*

The picture in Fig. 12 illustrates two retinal OCT scan views. The retinal structures appear in grayscale in the original image on the left. The Grad-CAM heatmap overlay, which is applied on the right, demonstrates the regions that the model centered on for classification. Warm colors emphasize the regions that are most significant for the model's predictions [14]. The heatmap displays how the model interprets important scan features, providing a visual explanation of the decision-making process.

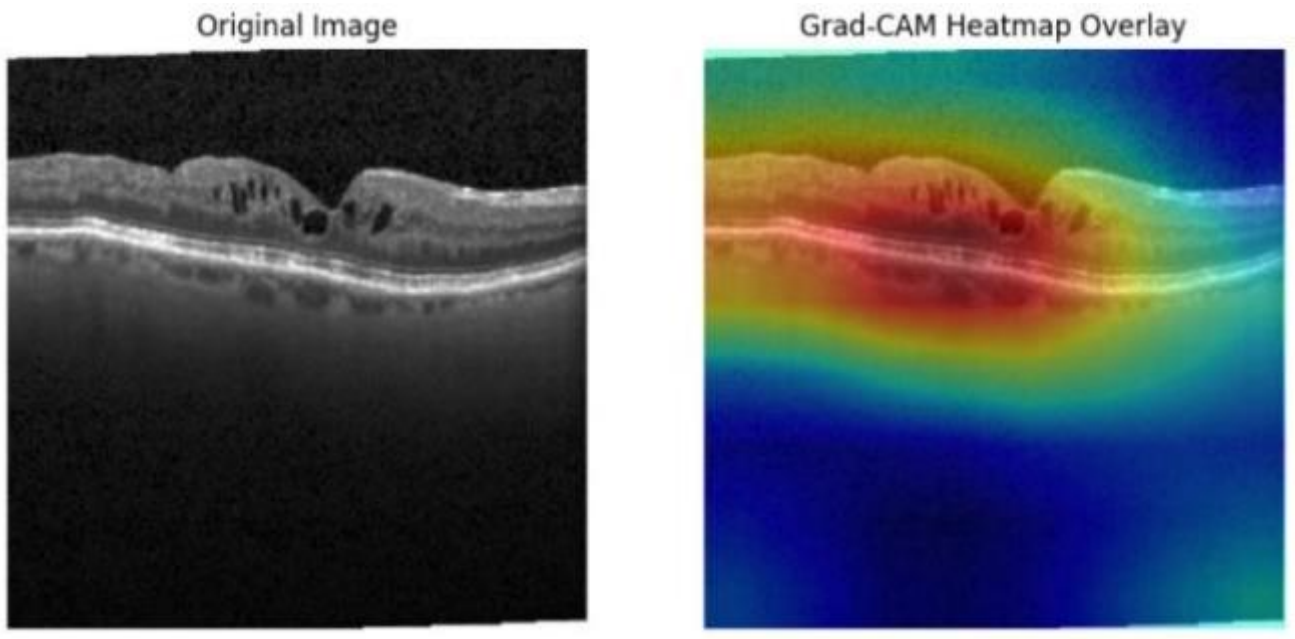

Fig. *12*. Anthracnose Identification using Grad-CAM

### *B. LIME*

The image in Fig. 13 indicates three LIME stages utilized in an OCT scan of the retina.

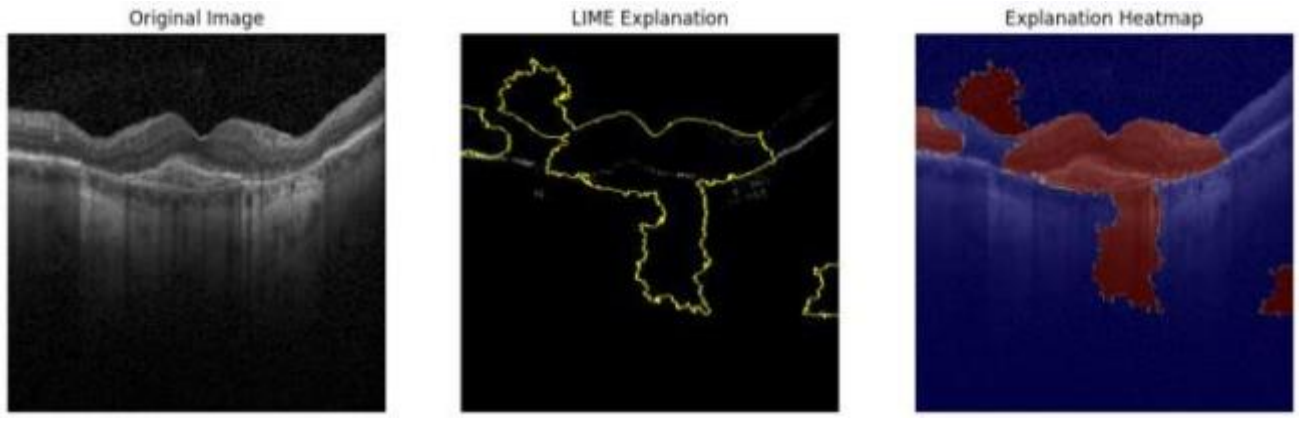

Fig. 13. Retinal Disease Identification using LIME

LIME emphasizes important areas with yellow boundaries in the middle of the original image on the left. The model's focus is clarified by a heatmap overlay on the right, which displays the most important features with darker red areas.

### *C. Occlusion Sensitivity*

Three stages of the occlusion sensitivity map for a retinal OCT scan are displayed in Fig 14.

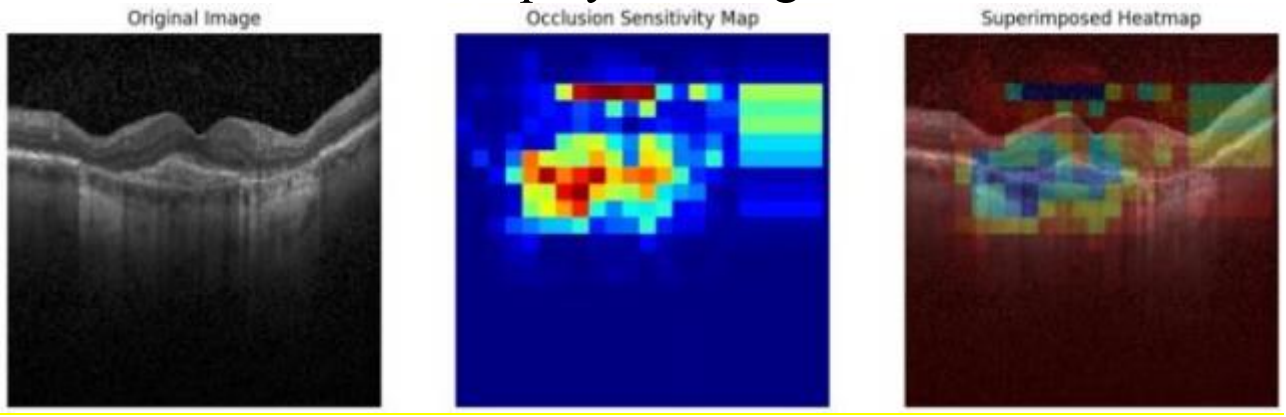

Fig. 14. Retinal Disease Identification using Occlusion Sensitivity

The retinal structure can be observed in the original image on the left. Bright colors signify high sensitivity, and the sensitivity map in the center highlights areas that are crucial for the model's decision. This map is superimposed over the heatmap on the right, which graphically depicts the important regions that go into the classification.

## VII. PROTOTYPE WEB APPLICATION

In order to deliver quick and dependable insights for early diagnosis and treatment, we developed an application shown in Fig. 15 that categorizes images of retinal diseases into distinct categories and provides accurate predictions alongside a confidence score.

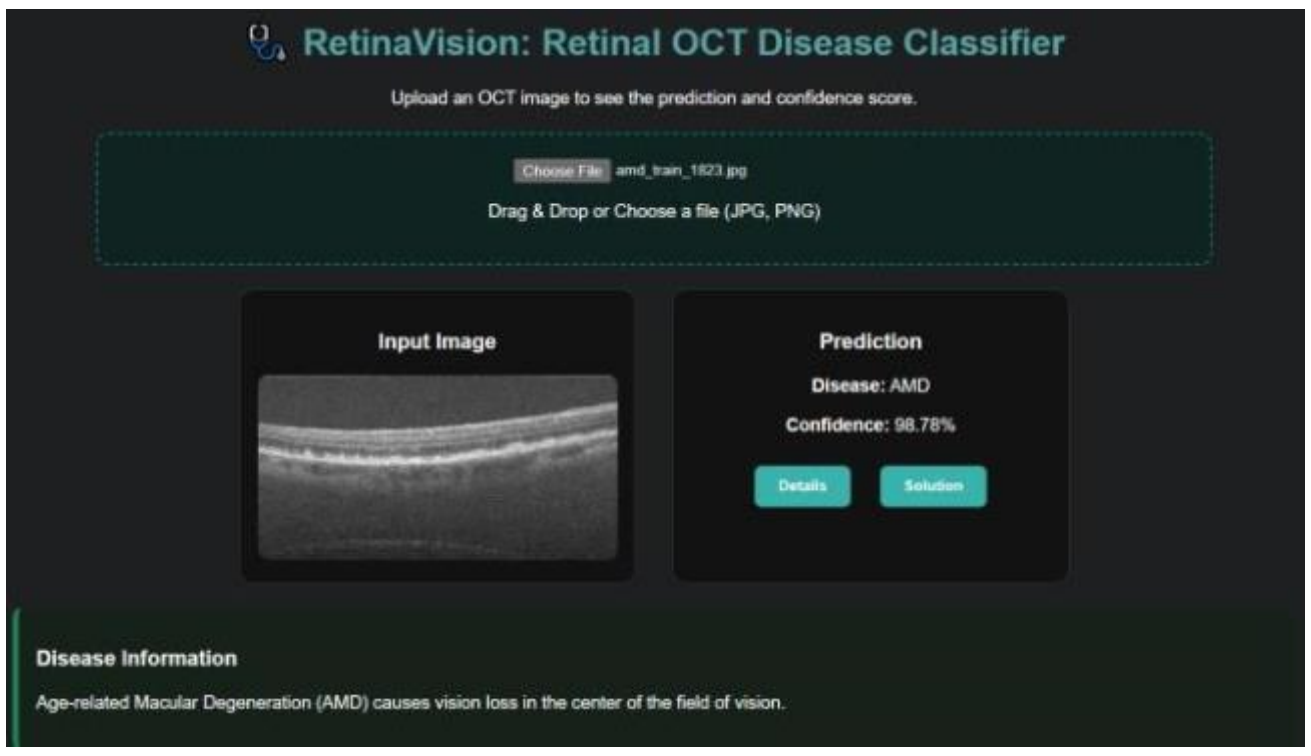

Fig. 15. Retinal OCT Disease Classifier Web Application

## VIII. COMPARISON

The performance of two algorithms, Xception and InceptionV3, is examined in Table 5. Xception reaches a 97.03% training accuracy and a 95.25% test accuracy. InceptionV3, in contrast, exhibits a slightly lower test accuracy of 94.82% but a slightly higher training accuracy of 97.83%.

TABLE V ACCURACY COMPARISON BETWEEN THE TWO ARCHITECTURES

| Algorithm | Xception | InceptionV3 |
|---|---|---|
| Training accuracy | 97.03% | 97.83% |
| Test accuracy | 95.25% | 94.82% |

The accuracy of the stated model is contrasted with previous studies in Table 6. Rithani et al [5] employ InceptionV3 to reach 92.76%, while Verma [4] utilized CNN to reach 84%. Eren et al. [7] applied transfer learning to achieve 91.47% whereas Vashanthi et al. used Nasnet to achieve 90%. RIDE helped Wali et al. [8] reach 80%. With 95.25%, the suggested Xception model performs better than the others, followed by InceptionV3 with 94.82%.

TABLE VII COMPARISON WITH PREVIOUS WORKS

| Author | Method | Accuracy (%) |
|---|---|---|
| Verma [4] | CNN | 84 |
| Rithani et al. [5] | InceptionV3 | 92.76 |
| Vasanthi et al. [6] | ResNet | 90 |
| Eren et al. [7] | Transfer Learning | 91.47 |
| Wali et al. [8] | RIDE | 80 |
| Proposed Model | Xception | 95.25 |
| | InceptionV3 | 94.82 |

## IX. CONCLUSION

This research offered a comprehensive framework for classifying retinal diseases from optical coherence tomography (OCT) images, utilizing deep learning approaches with more sophisticated architectures such as Xception and InceptionV3. The models trained and evaluated on the Retinal OCT Image Classification – C8 dataset achieved good accuracy across eight retinal diseases. Enhancements were made for generalization by means of augmentation methods such as CutMix and MixUp, while interpretability was addressed through Grad-CAM and LIME to produce visual explanations of model predictions to help build clinician trust. While the model performance was strong, the study also had limitations. The model has shown consistent performance, which suggest a reasonable robustness under typical non-ideal conditions. The dataset was a good size, but not comprehensive of all retinal diseases across various populations. Furthermore, the interpretability of the models for rare or borderline cases will necessitate further work. Nevertheless, the study highlights the promise of deep learning for automated retinal disease detection and describes exciting areas for further research and clinical translation.